\documentclass[conference]{IEEEtran}
\IEEEoverridecommandlockouts
\usepackage{cite}
\usepackage{amsmath,amssymb,amsfonts}
\usepackage{algorithm}
\usepackage{algpseudocode}
\usepackage{graphicx}
\usepackage{textcomp}
\usepackage{xcolor}
\usepackage{paralist}
\usepackage{caption}
\usepackage{subcaption}

\def\BibTeX{{\rm B\kern-.05em{\sc i\kern-.025em b}\kern-.08em
    T\kern-.1667em\lower.7ex\hbox{E}\kern-.125emX}}
\begin{document}
\makeatletter % changes the catcode of @ to 11
\newcommand{\linebreakand}{%
  \end{@IEEEauthorhalign}
  \hfill\mbox{}\par
  \mbox{}\hfill\begin{@IEEEauthorhalign}
}
\makeatother % changes the catcode of @ back to 12
\title{Vehicles Control: Collision Avoidance using Federated Deep Reinforcement Learning}
\UseRawInputEncoding
\author{\IEEEauthorblockN{Badr Ben Elallid$^1$, Amine Abouaomar$^2$, Nabil Benamar$^{1,2}$, and Abdellatif Kobbane$^3$}
\IEEEauthorblockA{$^1$ University Moulay Ismail, Meknes, Morocco. \\
$^2$ Al Akhawayn University in Ifrane, Morocco. \\
$^3$ ENSIAS, Mohammed V University in Rabat, Morocco.}
}

\maketitle
\begin{abstract}
In the face of growing urban populations and the escalating number of vehicles on the roads, managing transportation efficiently and ensuring safety have become critical challenges. To tackle these issues, the development of intelligent control systems for vehicles is paramount. This paper presents a comprehensive study on vehicle control for collision avoidance, leveraging the power of Federated Deep Reinforcement Learning (FDRL) techniques. Our main goal is to minimize travel delays and enhance the average speed of vehicles while prioritizing safety and preserving data privacy. To accomplish this, we conducted a comparative analysis between the local model, Deep Deterministic Policy Gradient (DDPG), and the global model, Federated Deep Deterministic Policy Gradient (FDDPG), to determine their effectiveness in optimizing vehicle control for collision avoidance. The results obtained indicate that the FDDPG algorithm outperforms DDPG in terms of effectively controlling vehicles and preventing collisions. Significantly, the FDDPG-based algorithm demonstrates substantial reductions in travel delays and notable improvements in average speed compared to the DDPG algorithm.
\end{abstract}
\section{Introduction}

The proliferation of vehicles on the urban roads led to increasing traffic congestion, longer travel times, and a higher number of road accidents \cite{elallid2022comprehensive,hanjri2023federated}. Legacy traffic management systems rely often on pre-determined rules, which may not be sufficient to tackle the continuous growing complexity and unpredictability of traffic scenarios. Recent work on machine learning opened new paths for developing intelligent vehicle control systems capable of adapting to the complex dynamic environments and making real-time decisions \cite{elallid2023reinforcement, elallid2022deep}. Deep Reinforcement Learning (DRL), as a subfield of machine learning, has demonstrated a high efficiency in solving complex control problems through enabling agents to acquire optimal policies by interacting with their environment \cite{feng2023dense, elallid2022dqn,kabbajdistfl}.

DRL combines the power of deep neural networks with reinforcement learning algorithms, enabling autonomous agents to learn optimal policies through trial and error in complex environments. By interacting with their surroundings, these agents can optimize their actions to achieve specific goals, such as minimizing travel time or avoiding collisions. In the realm of vehicle control, DRL has the potential to revolutionize how vehicles navigate through traffic, make driving decisions, and respond to unforeseen situations. By leveraging DRL techniques, we can develop advanced control systems that can autonomously adapt to dynamic traffic conditions, optimize driving behavior, and ultimately improve traffic efficiency and safety \cite{abouaomar2021deep,abouaomar2022federated}.

In the context of vehicle control, DRL can be leveraged to optimize driving decisions in order to avoid collision, route planning, and traffic efficiency.  we aim to go beyond the conventional use of DRL and explore the potential of Federated Deep Reinforcement Learning (FDRL) techniques, specifically the Federated Deep Deterministic Policy Gradient (FDDPG), for vehicle control in order to enhance collision avoidance and overall traffic management. One key aspect of our approach is the focus on privacy-preserving learning. We recognize the importance of keeping individual vehicle data private while still benefiting from the collective learning of other vehicles in their traffic environment. By utilizing FDRL techniques, we can achieve this objective, allowing vehicles to learn from each other's experiences without directly sharing sensitive data.

We provide also a comparison of local DDPG and global FDDPG models performance in terms of the travel delay and the average speed, also we assess their suitability for practical implementation in real-world traffic scenarios. Through extensive experimentation, we employed a widely-used traffic simulation frameworks, namely, Veins and SUMO, to evaluate the performance of the FDDPG algorithm for vehicle control and collision avoidance. We finally provided an analysis of our study to emphasis on the potential benefits and challenges associated with the adoption of FDRL techniques for vehicle control.

The contributions of this paper can be summararized as follows,
\begin{itemize}
    \item 
    We provided a comparison of the DDPG and FDDPG algorithms, then we evaluate their performance in terms of travel delay reduction and average speed improvement under different traffic scenarios.
    \item 
    We employed traffic simulation frameworks, namely, Veins and SUMO, to create a realistic environment for simulating vehicular networks and evaluating the effectiveness of the compared algorithms in real-world traffic setup.
    \item 
    We demonstrated the superiority of the FDDPG algorithm over the DDPG algorithm in controlling vehicles for collision avoidance and to improve traffic efficiency and safety.
\end{itemize}

The remainder of this paper is as follows. Section II discusses different related works. Section III presents the System model and different entities composing the considered model. Section IV discusses the proposed solutions and the comparison between FDDPG and DDPG setups. Section V is dedicated to discussing the finding. Section VI conclude the paper.

\section{Related Works}

The work of \cite{woo2020collision} developed a DRL-based collision avoidance method for unmanned surface vehicles (USVs) that focuses on the decision-making stage. The aim of this work is to determine if an avoidance maneuver is necessary and, if so, the direction of the maneuver. The authors proposed a neural network architecture and a semi-Markov decision process model for the USV collision avoidance. The DRL network is trained through a number of simulations and then implemented in experiments and simulations to evaluate its situation recognition and collision avoidance performance. The work in \cite{feng2021collision} aimed at exploring the potential of DRL into solving collision avoidance problems in unknown and compact environments. The proposed approach is compared to different traditional methods, such as potential field-based methods and the dynamic window approach. In \cite{raibail2022decentralized}, authors proposed a decentralized DRL framework for collision avoidance, where each agent independently makes decisions without communicating with other agents. The proposed approach enables mobile robot agents to learn efficient obstacle avoidance and navigation towards a targeted point in the environment with different dynamic obstacles. Soft actor-critic algorithm is employed to train the agents on obstacle avoidance policies in dynamic environments. In \cite{chang2021autonomous}, authors focused on enhancing the quality and safety of autonomous driving control by utilizing DRL as an alternative to traditional rule-based control strategies. Authors employed DDPG and Recurent DPG algorithms, combined with Convolutional Neural Networks (CNN), to enable autonomous driving control for self-driving cars.
In this paper, we aim at comparing most investigated methods from the literature and provide a comprehensive summary on pros and cons of each of the methods.

%------------- Brief introduction to DRL ---------------- ----------------------
\section{A brief overview of Federated Deep Reinforcement Learning}
DRL has emerged as a powerful technique in the field of Autonomous Vehicles (AVs) and has the potential to significantly improve the safety of autonomous driving. DRL is a lightweight technology that enables quick decisions and actions in real-time. DRL is based on the interaction between an agent and its environment, where the agent performs various actions and receives a reward for each performed action  \cite{palanisamy2018hands}. At each step $t$, the agent observes a state $S_{t}$ in the environment, chooses an action $a_{t}$ , gets a reward $r_{t+1}$, and the environment transitions to the next state $S_{t+1}$. One way to define Reinforcement learning is the Markov Decision Process (MDP), which includes several terms $(\mathcal{S},\mathcal{A},\mathcal{T},\mathcal{R})$ with the following conditions:
\begin{enumerate}
    \item $\mathcal{S}$ is a set of all possible states of an environment $(s \in \mathcal{S})$
    \item $\mathcal{A}$ is a set of actions that an agent can take $(a \in \mathcal{A})$
    \item $\mathcal{T}: \mathcal{S}\times \mathcal{A} \times \mathcal{S} \rightarrow [0,1]$ is the transition state that provides the probability associated with executing an action in a state and transitioning to a new state.
    \item $\mathcal{R}: \mathcal{S}\times \mathcal{A} \times \mathcal{S} \rightarrow \mathbb{R}$ is the reward that provides the penalty for taking an action in a state and transitioning to a new state.
\end{enumerate}
A probability of taking action $a_{t}$ in state $s_{t}$, so-called policy and denote it by $\pi$. This probability is expressed as follow:
\begin{equation*}
\pi(a_{t}|s_{t})= P[A=a_{t}|S=s_{t} ]
\end{equation*}
In contrast to (MDP), DRL is a model-free algorithm that does not require any probability modeling of the environment.

The DRL algorithm is suitable for use in the field of autonomous driving due to its capacity to manage new and unobserved environments. The DRL requires two main steps: 1) The agent tries to explore its environment by taking random actions; 2) The agent exploits prior knowledge by taking advantage of the exploration phase and follows an optimal policy $\pi_{*}$ to achieve the best action $a_{t}$ by maximizing the sum of subsequent rewards, identified as follows:
\begin{equation*}
G_{t}= r_{t} + \gamma r_{t+1} + \gamma^{2} r_{t+2} + ... + \gamma^{T} r_{t+T} =\sum\limits_{k=0}^T \gamma^{k} r_{t+k}
\end{equation*}

Where $\gamma\in[0, 1]$ represents the discount factor that penalizes future rewards, and T represents the end time of episode.
%Generally, we can classify DRL algorithms into two classes: (i) discrete algorithms such as DQN and DDQN; and (ii) continuous algorithms such as the stochastic policy technique (e.g., A3C) and the deterministic technique (e.g., DDPG). 
DDPG is an off-policy, model-free, actor-critic algorithm based on the Deterministic Policy Gradient (DPG) theorem \cite{silver2014deterministic}. 

%In contrast to Deep-Q-Learning-based methods, actor-critic policy-gradient-based methods can be easily applied to continuous action spaces. 

DDPG combines DQN and policy gradients to learn deterministic policies. DDPG learns Q-value and policy based on off-policy data and the Bellman equation \ref{eq}:

\begin{equation}
   \label{eq}
    Q_{*}(s_{t},a_{t}) = E \big[ r_{t} + \gamma \max_{a_{t+1}}Q_{*}(s_{t+1},a_{t+1}) \big]
\end{equation}
Federated Deep Reinforcement Learning (FDRL) combines Federated Learning and Deep Reinforcement Learning to enable collaborative and privacy-preserving training of reinforcement learning models. Multiple devices or entities, such as vehicles or IoT devices, train their own local models without sharing raw data. The central server aggregates model updates, creating a global model that captures collective knowledge. FDRL benefits from collaborative learning while maintaining data privacy and security. It has the potential to transform applications like autonomous vehicles, allowing vehicles to learn from each other without sharing sensitive data.

In this paper, we propose a technique for controlling autonomous vehicles (AVs) in traffic with the goal of reaching their destination safely and collision-free using Federated DDPG algorithm.
%======================= Figure of proposed method ====================
\begin{figure*}[htp]
    \centering
   \includegraphics[width=.6\linewidth,keepaspectratio]{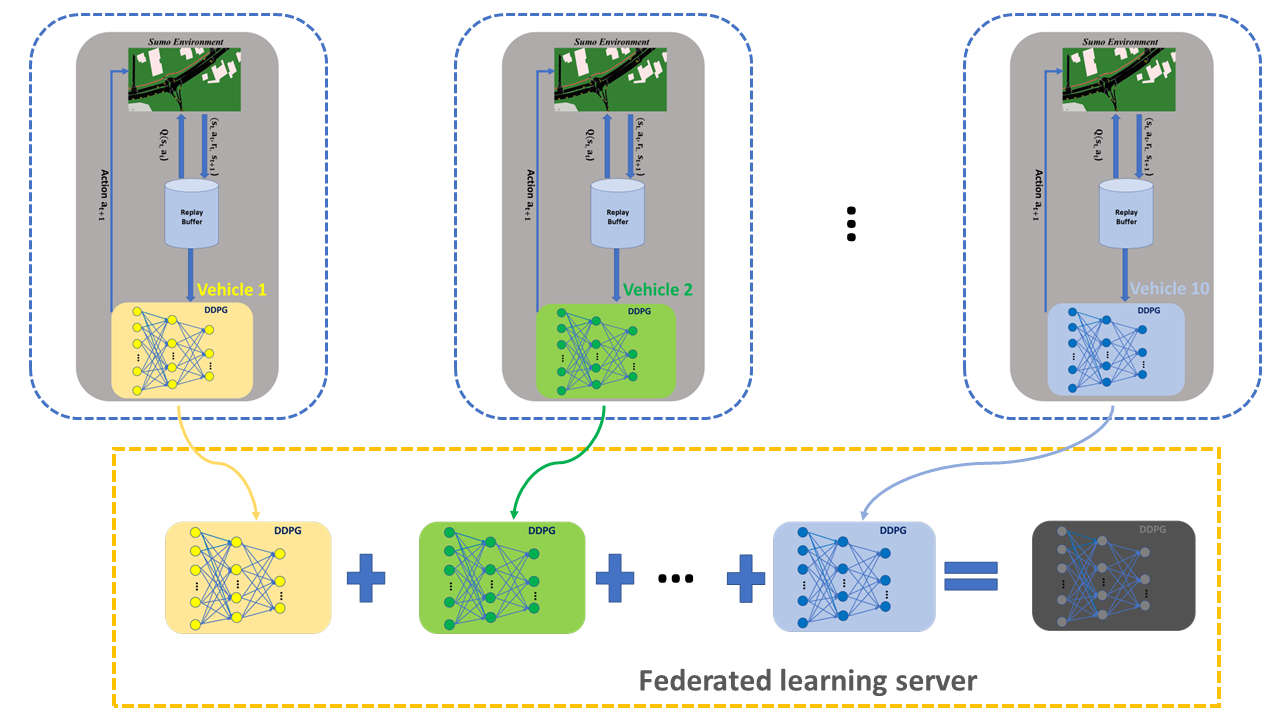}
    \caption{\small{The FDRL framework focuses on collision avoidance (RL) tasks in (AVs). In this framework, wireless communication is established between all ten agents and the federated learning (FL) server. Each agent operates autonomously within its own environment. Each round, the FL server collects and aggregates the RL models from all agents, generating a global model. This global model undergoes asynchronous updates from various RL agents, fostering collaborative learning within the system.}} 
    \label{fig:archi}
\end{figure*}   

\section{Federated Deep Reinforcement Learning}

\subsection{RL Problem Formulation}

\begin{compactitem}[$\bullet$]
    \item \textbf{State space}: Our simulation consists of various features of the vehicle received by the environment in each step. Specifically, the state at time step $t$ is represented by a vector $S_t = (PosX_t, PosY_t, Speed_t, \theta_t, A_t, Dest)$, where $PosX_t$ and $PosY_t$ denote the x and y positions of the vehicle, $\theta_t$ is its orientation, $Speed_t$ and $A_t$ are its velocity and acceleration, respectively. Additionally, $Dest$ represents the distance to the destination, which is calculated as the Euclidean distance between the current position of the vehicle $Pos_{AV}$ and the end position $Pos_{end}$, i.e., $|| Pos_{AV} - Pos_{end}||$. To obtain these features, we use TraCi, which is a tool provided by Sumo Simulator as shown in Figure \ref{fig:archi}.

    \item \textbf{Action space}: is a control command that is sent to our vehicle in the environment. Since DDPG is a continuous DRL algorithm, it is necessary to use continuous action. In our case, we use acceleration or deceleration as our control command that we send to the vehicle in the Sumo Simulator.

    %DQN is a discrete algorithm that utilizes five actions: $-1$, $1$, $2$, $5$, and $8$. Negative numbers indicate deceleration while positive numbers indicate acceleration.
    \item \textbf{State Transition}: is the probability of executing action $a_{t}$ in state $s_{t}$ at time $t$ and transitioning new state $s_{t+1}$ at time $t+1$:
        $P = P(s_{t+1}|s_{t}, a_{t})$
    \item \textbf{Reward function}: is a function that generates reward $r_{t}$ while executing action $a_{t}$. In our case, the vehicle should reach its destination as quickly as possible without colliding with other vehicles on its route, reducing time spent waiting at traffic lights or if some vehicles are braking. Let $V_{Speed}$, $Pos$, $Pos_{dest}$, $Traffic_{braking}$, $Traffic_{waiting}$ denote the speed of vehicles, the vehicle's position,the destination's position, and our vehicle is braking or waiting in traffic light, respectively. If the vehicle's speed is not zero, then a penalty of $0.04$ is assessed. If the simulator detects a crash, a penalty of $-5$ is obtained. To reduce waiting time, the reward is $-0.05$ if our vehicle brakes and waits at a traffic light. The reward is $-0.025$ if our vehicle is braking or waiting at a traffic light; otherwise, the penalty of $-0.02$ is attributed to the agent. Furthermore, the agent’s reward of $5$ is affected if the vehicle reaches its target destination.
    \begin{equation*}
reward = \left\{
    \begin{array}{llll}
        -10 & \text{if} \hspace{0.1cm} \text{there} \hspace{0.1cm} \text{is} \hspace{0.1cm} \text{collision} \\
        0.04 & \text{if} \hspace{0.1cm}  V_{Speed} \ne 0\\
        -0.05 & Traffic_{braking} \hspace{0.1cm} \text{and} \hspace{0.1cm} Traffic_{waiting} \\
        0.025 & Traffic_{braking} \hspace{0.1cm} \text{or} \hspace{0.1cm} Traffic_{waiting} \\
         0.05 & \text{not} \hspace{0.1cm} Traffic_{braking} \hspace{0.1cm} \\
         & \text{or} \hspace{0.1cm} \text{not} \hspace{0.1cm} Traffic_{waiting} \\
        10 & \text{To} \hspace{0.1cm} \text{reach} 
        \hspace{0.1cm} \text{destination} \\
        -0.02 & \text{by} \hspace{0.1cm} \text{default}
    \end{array}
\right.
\end{equation*}
    
\end{compactitem}
\vspace{\baselineskip}
\subsection{Federated DDPG model}

 %   \item \textbf{DQN model} : The training process for our model involves several steps. To start, we initialize the network with random weights. The agent then takes actions, which are either exploration or exploitation, and records them in a replay buffer along with the corresponding state $s_{t}$, reward $r_{t}$, and next state $s_{t+1}$, as a tuple $(s_{t},a_{t},r_{t},s_{t+1})$. We randomly select a batch of size 256 from the replay buffer, which is passed to the policy network to predict the Q-value. To update the Q-value using Equation 1, we perform a second pass through the network to calculate the maximum Q-value of the next action $a_{t+1}$. This second pass is referred to as the target network. We then calculate the loss between the predicted Q-value and the target Q-value using mean square error (MSE) and the ADAM optimizer, and update the weights of the network using backpropagation.
    
Our DDPG model consists of two networks: the Actor and the Critic. The Critic predicts the action value based on the state and the value obtained by the Actor. The Actor network takes the state $s_{t}$ as input and predicts the action taken by the agent. Ornstein-Uhlenbeck noise is added to this action to encourage exploration, as part of the policy ($a_{t} = \mu(s_{t}|\theta^{\mu}) + N_{t}$). The agent's experience, including the state $s_{t}$, action taken $a_{t}$, reward received $r_{t}$, and next state $s_{t+1}$, is stored as a tuple in the replay buffer. We randomly select a batch of tuples from the replay buffer, and the Critic network takes the state $s_{t}$ and action taken $a_{t}$ to predict the Q-value. To update the Q-value using Equation 1, we need a second pass through both the Actor and the Critic networks, which are referred to as the target Actor and the target Critic networks, respectively. Next, we compute the loss between the predicted Q-value and the value obtained from Equation \ref{eq}. Finally, we update the weights of the Actor and Critic networks using the following formulas: $\theta^{'} \longleftarrow  \tau \theta + (1 - \tau)\theta^{'}$ and $\phi^{'} \longleftarrow  \tau \phi + (1 - \tau)\phi^{'}$. Here, $\theta$ and $\phi$ are the weights of the Actor and Critic networks, respectively, and $\tau$ is a hyperparameter that controls the rate of updating the target networks.

The training process of Federated DDPG consists of five rounds. In each round, ten agents representing ten vehicles are trained in their respective traffic environments. After training, the weights of each agent are sent to an aggregation process to calculate the average of weights using this formula. 
\begin{equation*}
    w_{avg} = \frac {\sum_{i=1}^{N} n_i .w_i}  {\sum_{i=1}^{N} n_i}
\end{equation*}
Where $w_i$ represents the weights of agent $i$, $N$ represents the total number of agents, and $n_i$ is is the number of episodes contributed by agent $i$ 

In order to illustrate more how the FDDPG works, we provide the following pseudo-code
\begin{algorithm}[htp]
  \caption{Federated Deep Deterministic Policy Gradient (FDDPG)}
  \begin{algorithmic}[1]
    \State Initialize global critic network $Q_{\phi}(s, a)$ with random weights $\phi$
    \State Initialize global actor network $\mu_{\theta}(s)$ with random weights $\theta$
    \State Initialize target critic network $Q'_{\phi}(s, a)$ with weights $\phi' \leftarrow \phi$
    \State Initialize target actor network $\mu'_{\theta}(s)$ with weights $\theta' \leftarrow \theta$

    \For{each federated device $i$ in parallel}
        \State Initialize local critic network $Q_{\phi_i}(s, a)$ with weights $\phi_i \leftarrow \phi$
        \State Initialize local actor network $\mu_{\theta_i}(s)$ with weights $\theta_i \leftarrow \theta$
        \State Initialize local replay buffer $\mathcal{D}_i$
    \EndFor

    \For{each iteration}
        \For{each federated agent $i$ in parallel}
            \State Sample mini-batch of experiences from local replay buffer $\mathcal{D}_i$
            \State Update local critic network $Q_{\phi_i}(s, a)$ by minimizing the loss function:
            \State \hspace{\algorithmicindent} $L(\phi_i) = \frac{1}{N}\sum_{j=1}^{N}(Q_{\phi_i}(s_j, a_j) - y_j)^2$
            \State Update local actor network $\mu_{\theta_i}(s)$ using the policy gradient:
            \State \hspace{\algorithmicindent} $\nabla_{\theta_i}J(\theta_i) \approx \frac{1}{N}\sum_{j=1}^{N}\nabla_aQ_{\phi_i}(s_j, a_j)\nabla_{\theta_i}\mu_{\theta_i}(s_j)$
            \State Update local target networks:
            \State \hspace{\algorithmicindent} $\phi_i' \leftarrow \tau \phi_i + (1 - \tau) \phi_i'$
            \State \hspace{\algorithmicindent} $\theta_i' \leftarrow \tau \theta_i + (1 - \tau) \theta_i'$
        \EndFor

        \State Aggregate local critic network weights $\{\phi_i\}$ and actor network weights $\{\theta_i\}$ at the global server
        \State Update global critic network $Q_{\phi}(s, a)$ with aggregated weights
        \State Update global actor network $\mu_{\theta}(s)$ with aggregated weights
        \State Update global target networks:
        \State \hspace{\algorithmicindent} $\phi' \leftarrow \tau \phi + (1 - \tau) \phi'$
        \State \hspace{\algorithmicindent} $\theta' \leftarrow \tau \theta + (1 - \tau) \theta'$
    \EndFor
  \end{algorithmic}
\end{algorithm}

\section{Simulation Results}
\subsection{Setup}
%\begin{figure}[htp]
%    \centering
%   \includegraphics[width=\columnwidth,keepaspectratio]{map.png}
%    \caption{Illustration of our urban scenario on the Meknes city map. } 
%    \label{fig:map}
%\end{figure}
\begin{figure}%
    \centering
    \subfloat[\centering  ]{{\includegraphics[width=6.5cm,keepaspectratio]{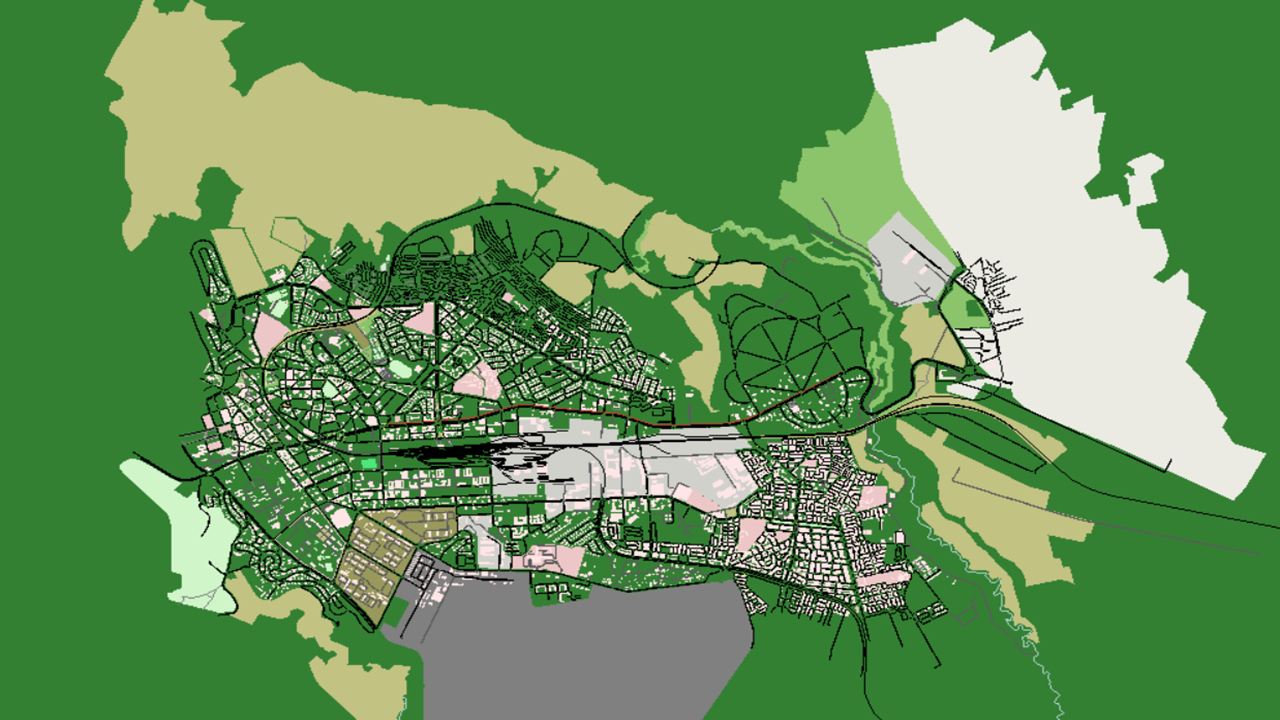} }}%
    \qquad
    \subfloat[\centering  ]{{\includegraphics[width=6.5cm,keepaspectratio]{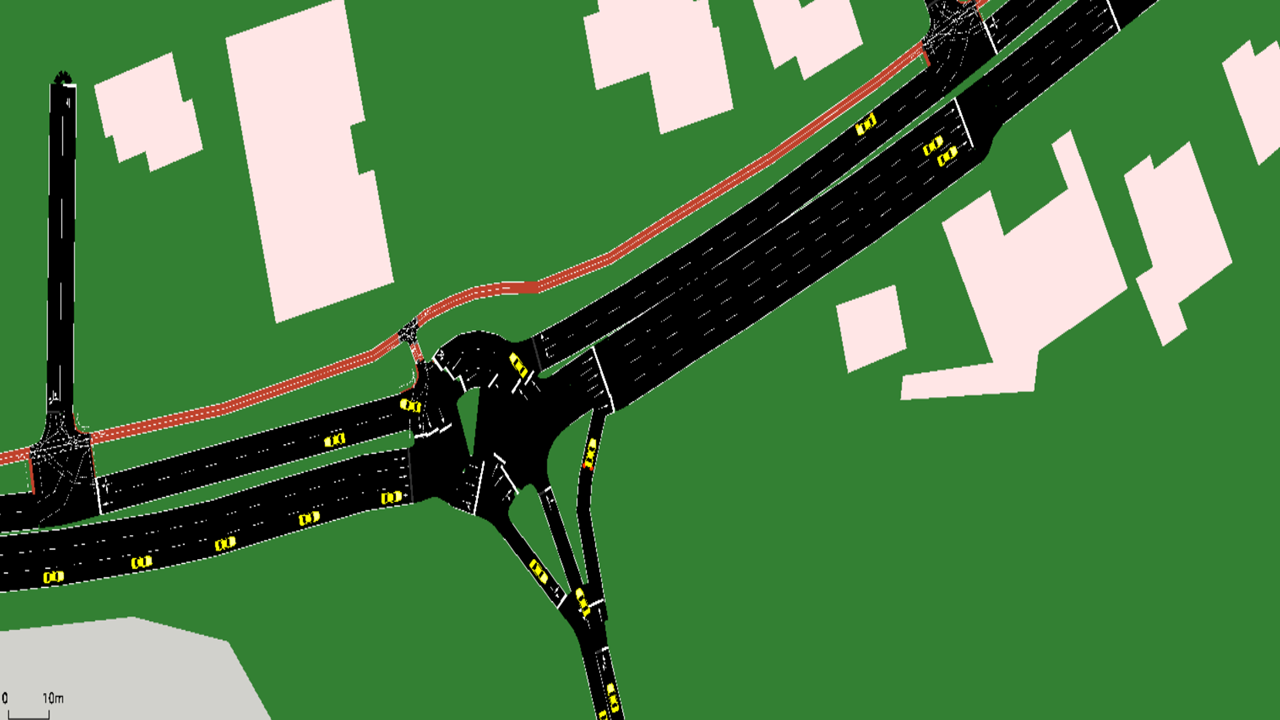} }}%
    \caption{\small{Illustration of Meknes city map which used like our environment to train FDDPG models. Figure (a) illustrates the hole map used in our training, while Figure (b) shows an example of traffic in our map involving other vehicles.}}%
    \label{fig:map}%
\end{figure}
Nowadays, testing autonomous vehicle approaches in hyper-realistic virtual environments is the most important part of discovering comfortable AV systems. According to the complexity of the scenarios of autonomous driving, it is necessary to simulate and validate the proposed approaches in simulators close to the real world in order to ensure the efficiency of these models. For these reasons, we evaluate our model in Sumo Simulator \cite{SUMO2018}, which is considered the most powerful simulator for traffic simulation since it allows export of maps from the real world using Open Street Maps. Open Street Maps allows us to import a large road network from maps, which allows us to simulate our model in real-world conditions. The PythonAPI is provided by SUMO and manages various scenarios. Further, we combine Sumo with Gym, which is an open-source Python library created by OpenAI \cite{brockman2016openai}. It was initially intended to be used for the creation and evaluation of reinforcement learning algorithms. The gym facilitates communication between the DRL algorithm and the environment. For this purpose, our model aims to benefit from the power of OpenAI and Sumo simulation to accelerate the training of our model and get the best results. On the other hand, we use TraCi \cite{wegener2008traci} to control and communicate with vehicles in our environment.

%After NB
We imported a map of the city of Meknes, Morocco using Open Street Map, as shown in Figure \ref{fig:map}, to simulate 4868 vehicles navigating randomly across the map. The speed limit and the collision sensitivity were set to 20 $m/s$ and 2 respectively.  

The training episodes terminate under one of the following conditions: 1) the ego vehicle collides with other vehicles, 2) the ego vehicle reaches its destination, or 3) the number of steps in the episode exceeds the maximum steps (900).
We have designed our architecture with two fully connected hidden layers in both the actor and critic networks. The actor network consists of 400 neurons in the first hidden layer and 300 neurons in the second hidden layer. Similarly, the critic network also has 400 neurons in the first hidden layer and 300 neurons in the second hidden layer. The input layer is of size 6, representing the shape of the state $S_{t}$, and the output layer size is 1, predicting the action. To prevent the issue of vanishing gradients during backpropagation, we used the rectified linear unit as our activation function. We implemented our proposed model in Python, using the PyTorch library to construct the DDPG architecture. Training was conducted on a machine equipped with an Intel Core i7-11800H (8-Core, 2.3GHz) processor, 16GB of RAM, and a single NVIDIA RTX 3050 GPU. We employed the Adam optimizer, and additional parameters are specified in Table \ref{tab1}.
\begin{table}[htp]
 \centering
\caption{Parameters of DDPG model}
\label{tab1}
\begin{tabular}{ p{3cm}p{3cm}}
\hline
\textbf{Parameter}& \textbf{Value} \\ \hline
Actor/critic learning rate & $5\times 10^{-4}$ \\ \hline
Episodes & 500 for each agent \\ \hline
Actor/critic batch size & 64 \\ \hline
$\gamma$ &  0.99 \\  \hline
Replay Memory Size & 50000 \\ \hline
\end{tabular}
\end{table}
\subsection{Result}
\begin{figure}[htp]
  \includegraphics[width=7.0cm,keepaspectratio]{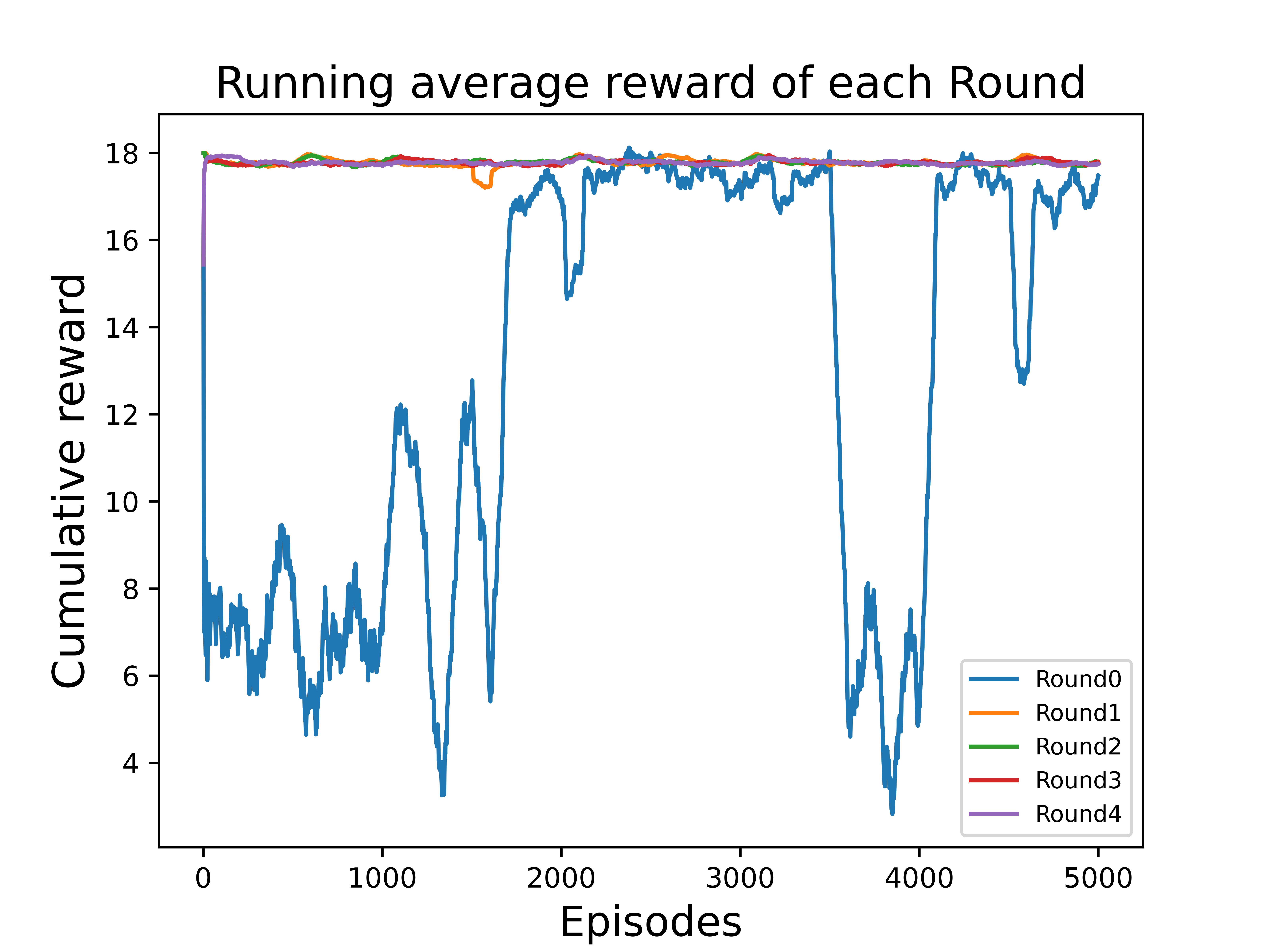}
  \caption{Average reward of each round during episodes}\label{fig:av_reward}
\end{figure}
The average rewards of the FDDPG model are shown in Figure \ref{fig:av_reward}. In the initial round, the curve exhibits fluctuations, reaching both high and low values. This variability arises from variations in learning among the agents, with some learning faster than others. However, starting from the second round, we observe a consistent increase in the average reward, eventually stabilizing. Notably, in the final round, all agents achieve a high and stable score, indicating the rapid learning capability of our model.

In our evaluation, we aim to determine the effectiveness of two models in controlling vehicles within their traffic scenarios. We analyze multiple metrics including collisions, travel delay, and average vehicle speed to assess their performance. To ensure accurate and reliable results, we conduct 20 simulation episodes and measure the aforementioned metrics.

During the simulation, the AVs are controlled by either a trained actor network for the local DDPG model or the global FDDPG model. The objective is to guide the AVs towards their respective target destinations while minimizing travel delay and avoiding collisions. To facilitate a comprehensive evaluation, we specifically chose four AVs and conducted performance comparisons between the local model DDPG and the global model FDDPG.

\begin{compactitem}[$\bullet$]

\item \textbf{Collision}: We assessed the performance of an Autonomous Vehicle (AV) using Traci and FDDPG controllers through simulation. Specifically, we focused on measuring the collision occurrences at various destination positions. We chose five destination positions located at Euclidean distances of $10 m$, $20 m$, $52 m$, $107 m$, and $207 m$ from the starting point to the endpoint. Our findings indicate that when the AV was controlled by the FDDPG controller, it effectively avoided collisions with other vehicles. Conversely, when the AV was controlled by DDPG, collisions with other vehicles were observed in every episode.

\item \textbf{Travel delay} : we evaluated the impact of local and global models on reducing travel delay for four AVs across five different distances. Figure \ref{fig:Travel} provides a  illustrated of travel delay for AVs. Our findings revealed that both the local and global models contributed to a significant reduction in travel delay, allowing the AVs to reach their destinations more efficiently. Notably, the global model, specifically the FDDPG model, outperformed the local DDPG model in terms of facilitating faster achievement of the AVs' destinations.

\item \textbf{Average speed}: To evaluate the speed performance of the autonomous vehicles (AVs), we analyzed their average speed across five different distances. The average speed was computed by summing the speeds of the AVs over a single episode and dividing it by the number of steps taken to reach their destinations. Figure \ref{fig:speed} provides a visual representation of the results.

Our analysis revealed that the average speed of the AVs, controlled by both the DDPG and FDDPG algorithms, increased as the distance between the start and end points increased. This suggests that the AVs learned to accelerate their speeds to reach their destinations more quickly. Additionally, we observed that the FDDPG algorithm outperformed the DDPG algorithm in terms of achieving higher average speeds for the four AVs.

\end{compactitem}
\vspace{\baselineskip}
 \begin{figure*}
     \centering
     \begin{subfigure}[b]{0.22\textwidth}
         \centering
         \includegraphics[width=\textwidth,keepaspectratio]{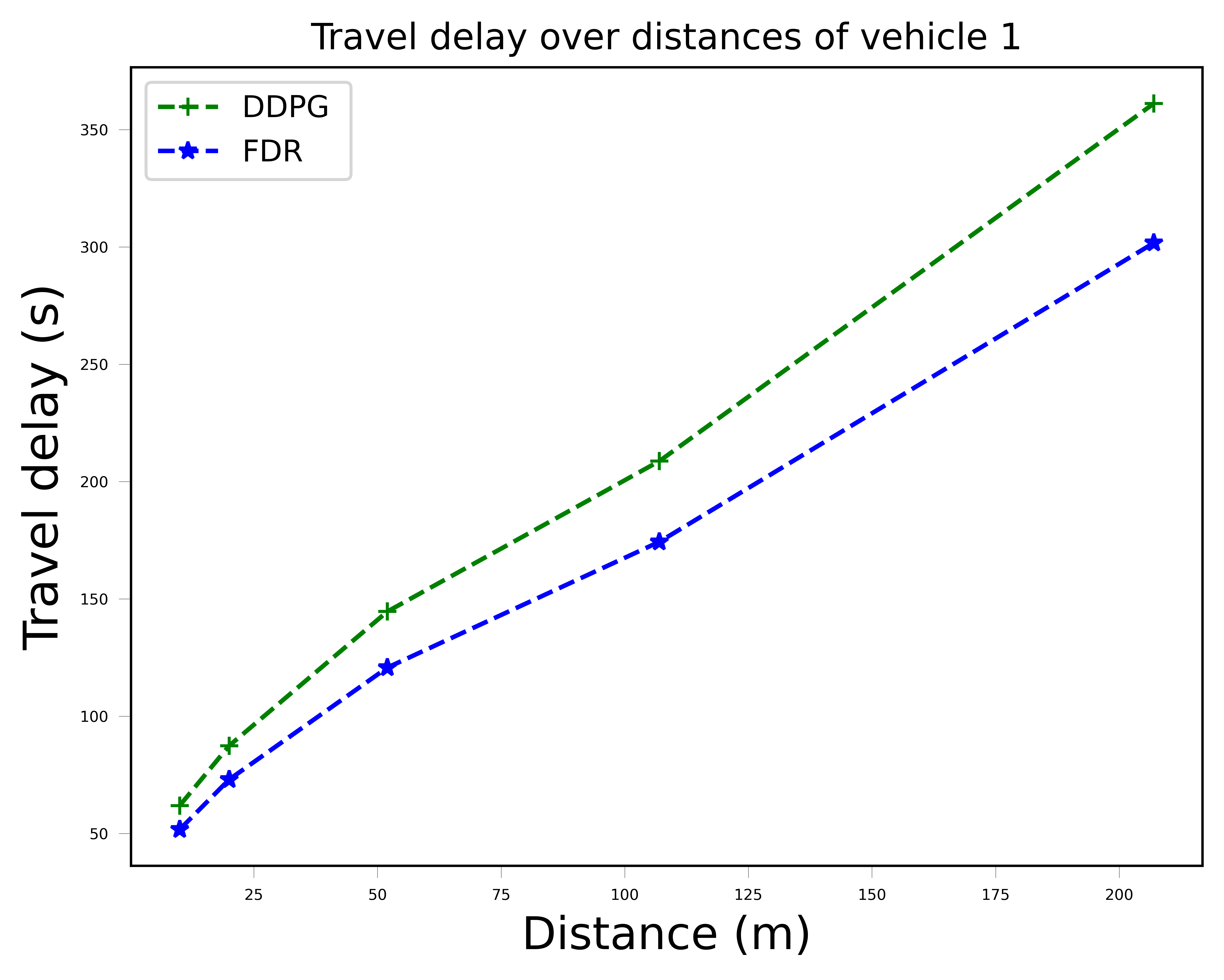}
         \caption{$Vehicle_1$}
         \label{fig:vehicle1}
     \end{subfigure}
     \hfill
     \begin{subfigure}[b]{0.22\textwidth}
         \centering
         \includegraphics[width=\textwidth,keepaspectratio]{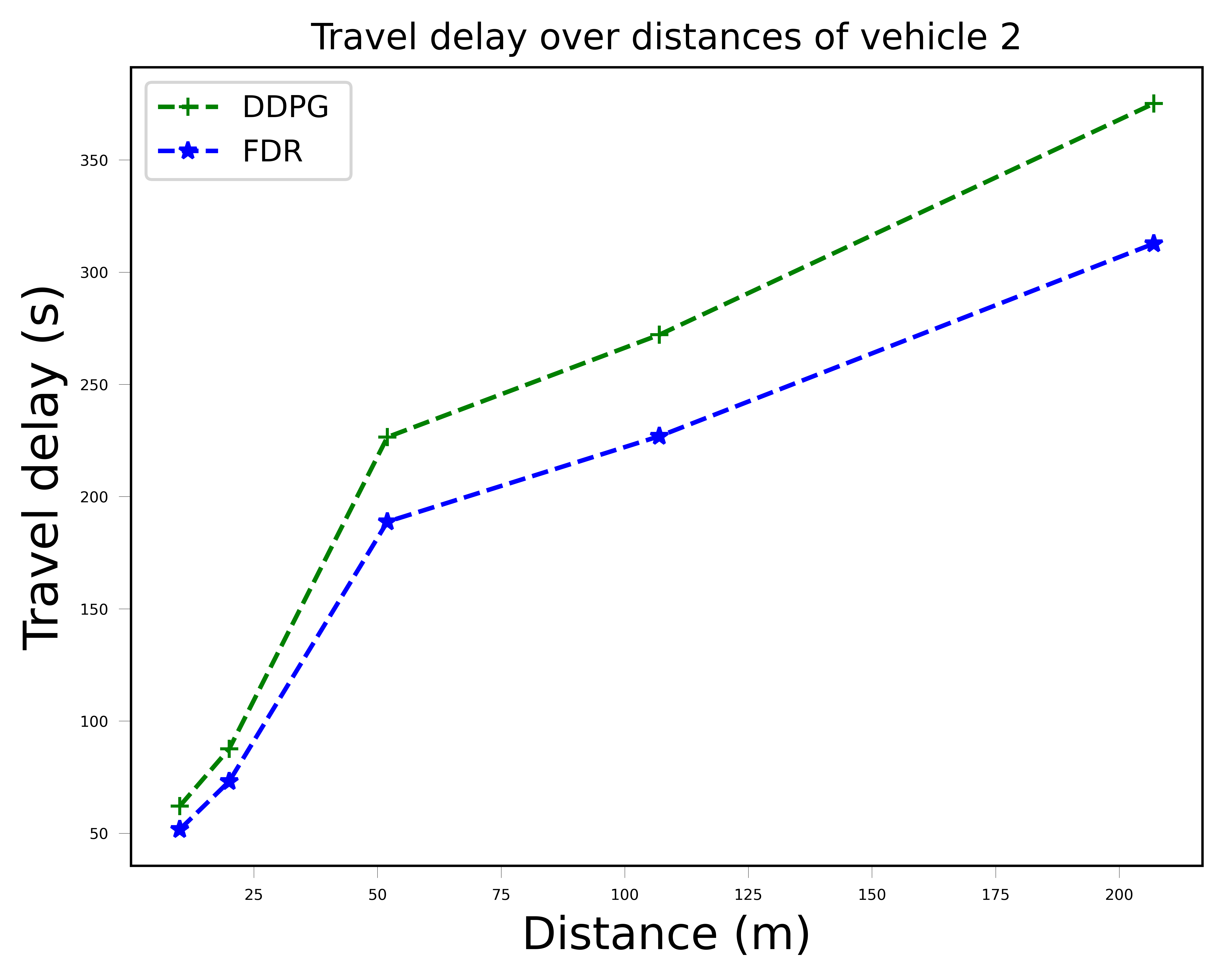}
         \caption{$Vehicle_2$}
         \label{fig:vehicle2}
     \end{subfigure}
     \hfill
     \begin{subfigure}[b]{0.22\textwidth}
         \centering
         \includegraphics[width=\textwidth,keepaspectratio]{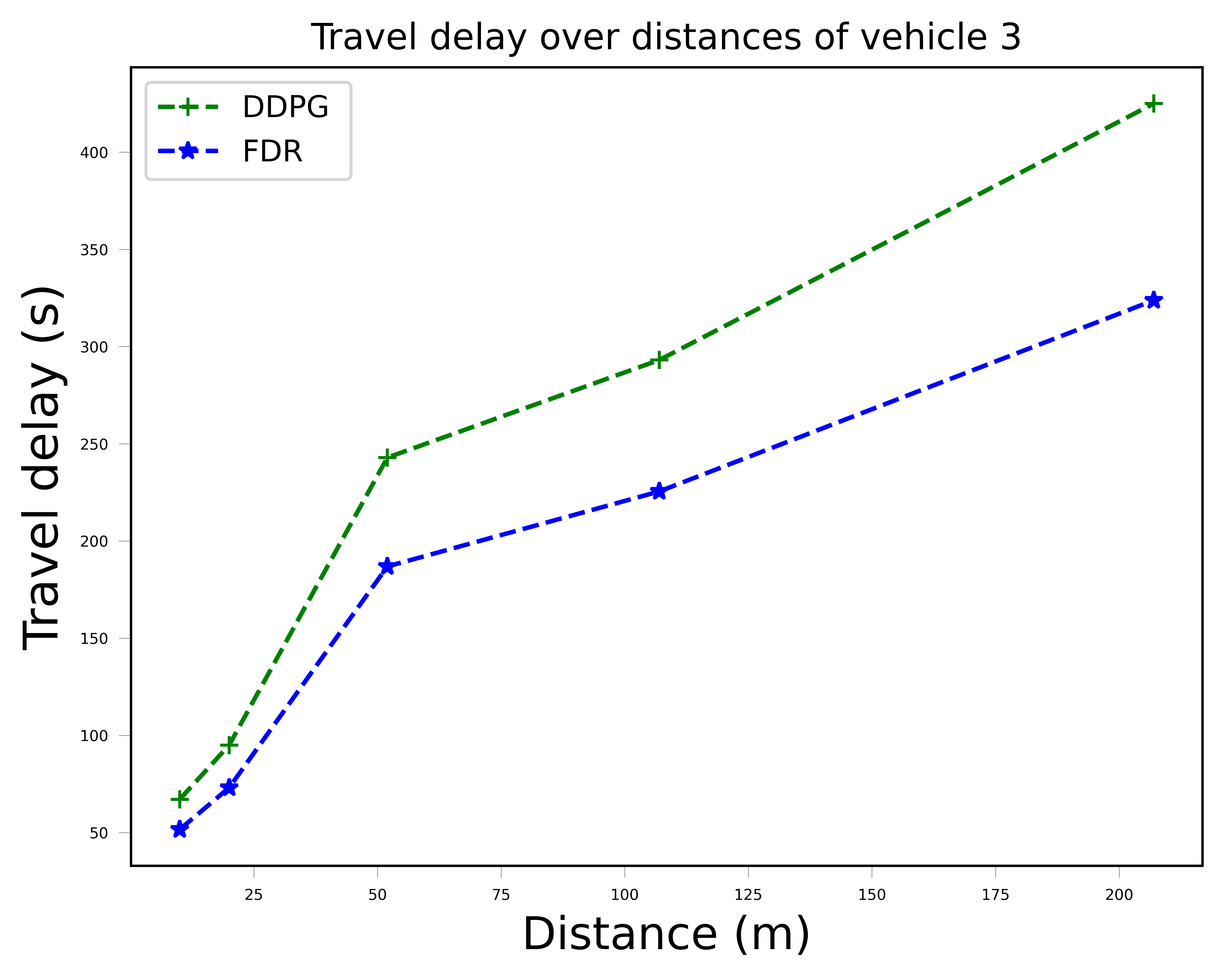}
         \caption{$Vehicle_3$}
         \label{fig:vehicle3}
     \end{subfigure}
     \begin{subfigure}[b]{0.22\textwidth}
         \centering
         \includegraphics[width=\textwidth,keepaspectratio]{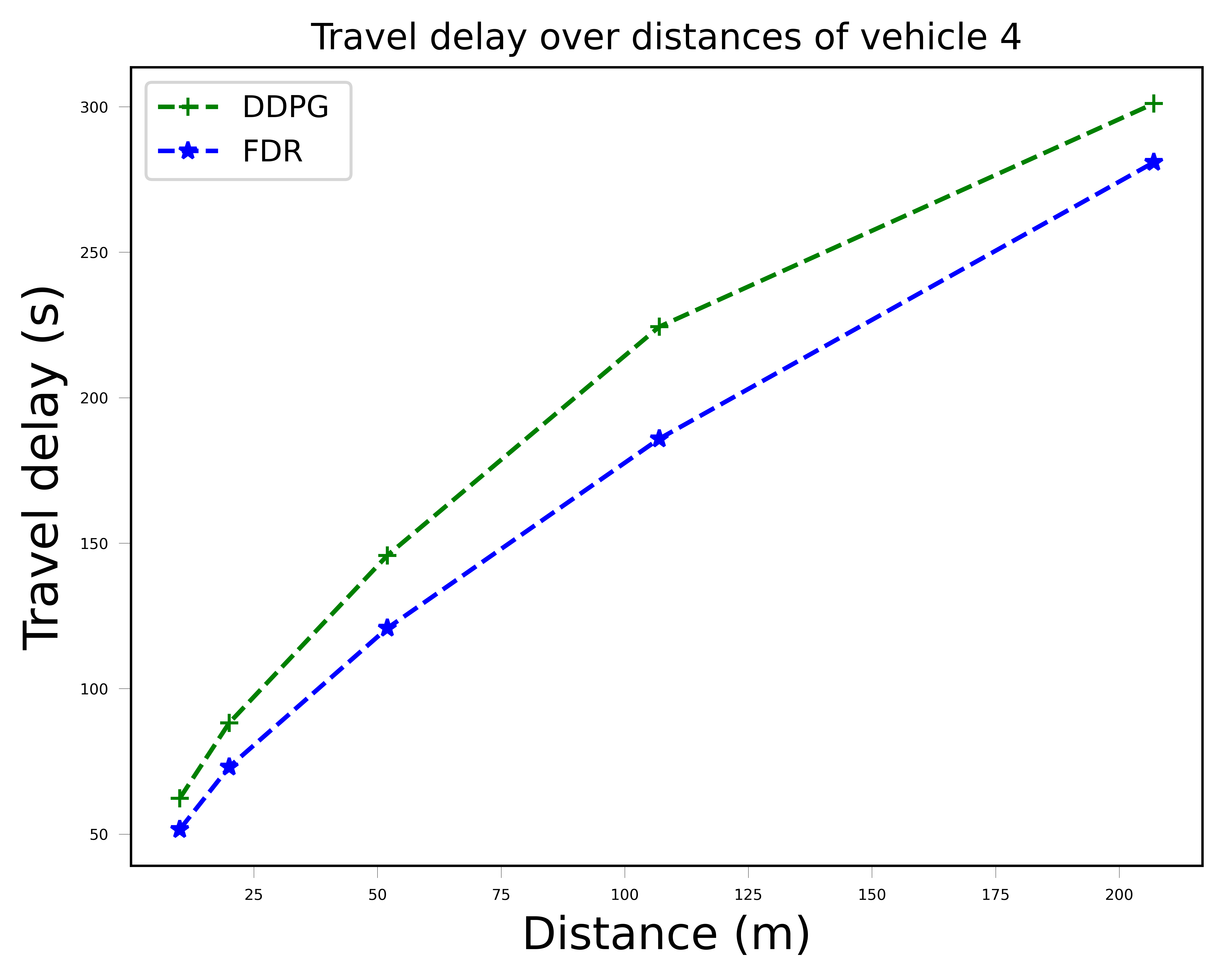}
         \caption{$Vehicle_4$}
         \label{fig:vehicle4}
     \end{subfigure}
        \caption{Travel delay of four vehicles in different distances }
        \label{fig:Travel}
\end{figure*}
\begin{figure*}
     \centering
     \begin{subfigure}[b]{0.22\textwidth}
         \centering
         \includegraphics[width=\textwidth,keepaspectratio]{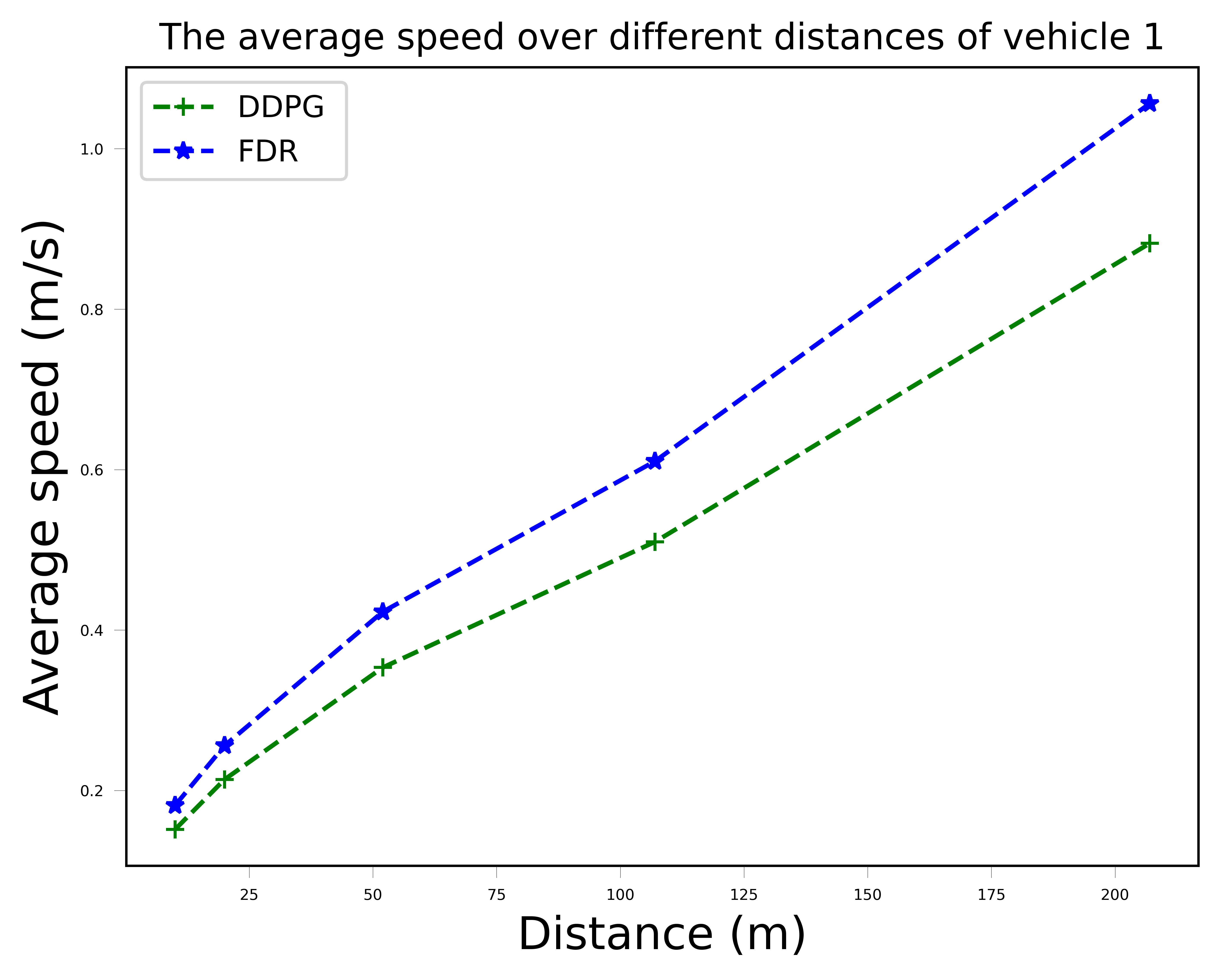}
         \caption{$Vehicle_1$}
         \label{fig:speed1}
     \end{subfigure}
     \hfill
     \begin{subfigure}[b]{0.22\textwidth}
         \centering
         \includegraphics[width=\textwidth,keepaspectratio]{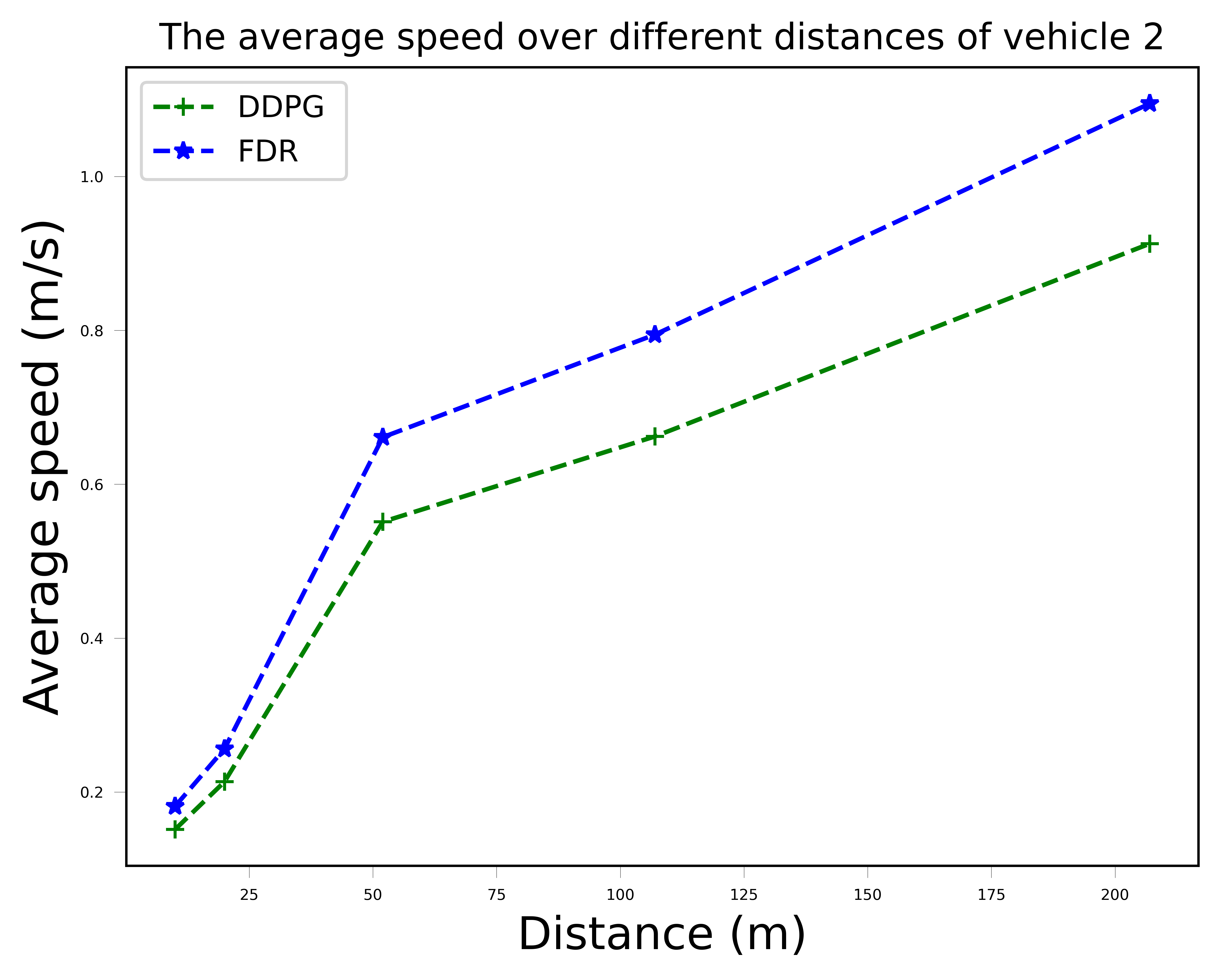}
         \caption{$Vehicle_2$}
         \label{fig:speed2}
     \end{subfigure}
     \hfill
     \begin{subfigure}[b]{0.22\textwidth}
         \centering
         \includegraphics[width=\textwidth,keepaspectratio]{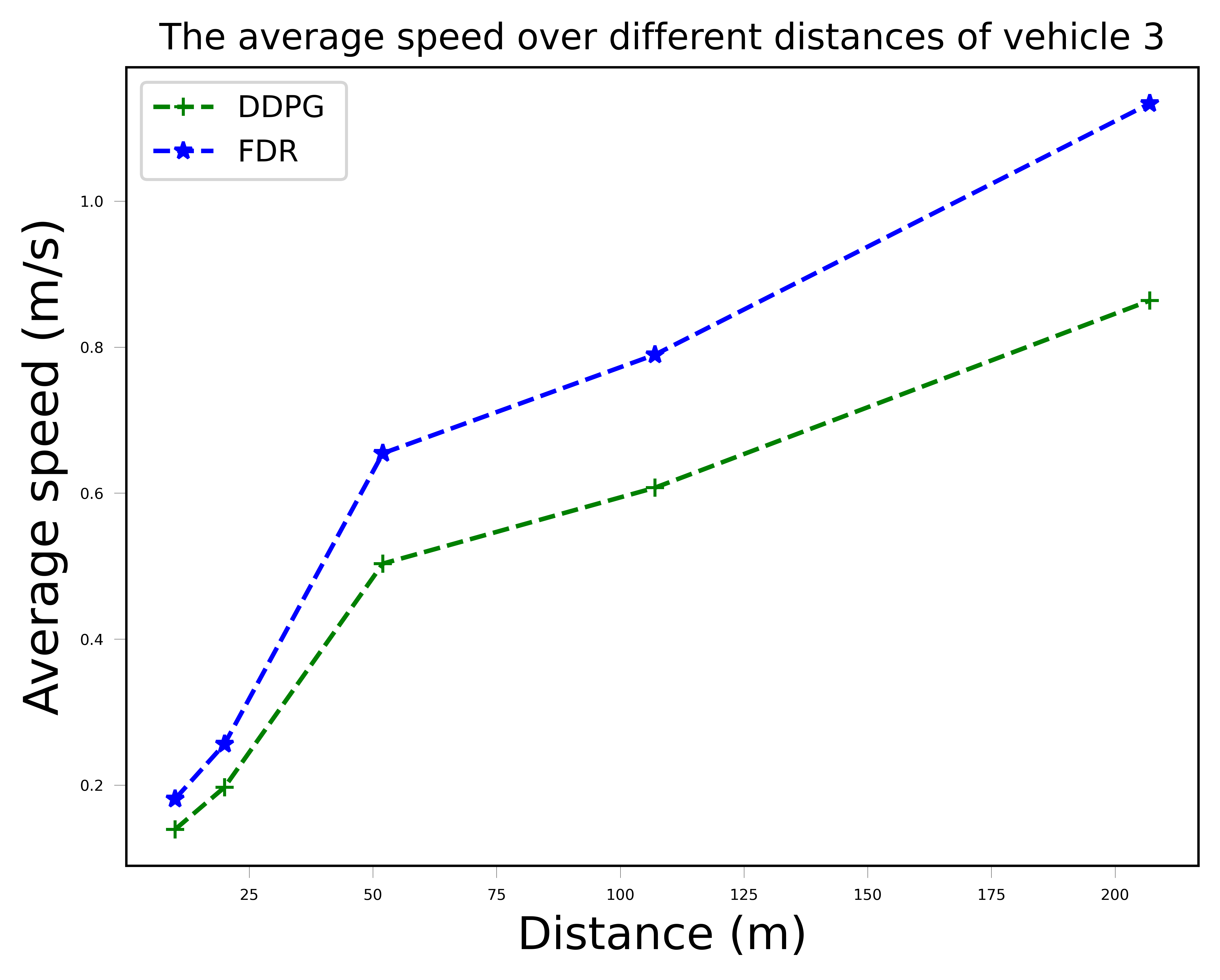}
         \caption{$Vehicle_3$}
         \label{fig:speed3}
     \end{subfigure}
     \begin{subfigure}[b]{0.22\textwidth}
         \centering
         \includegraphics[width=\textwidth,keepaspectratio]{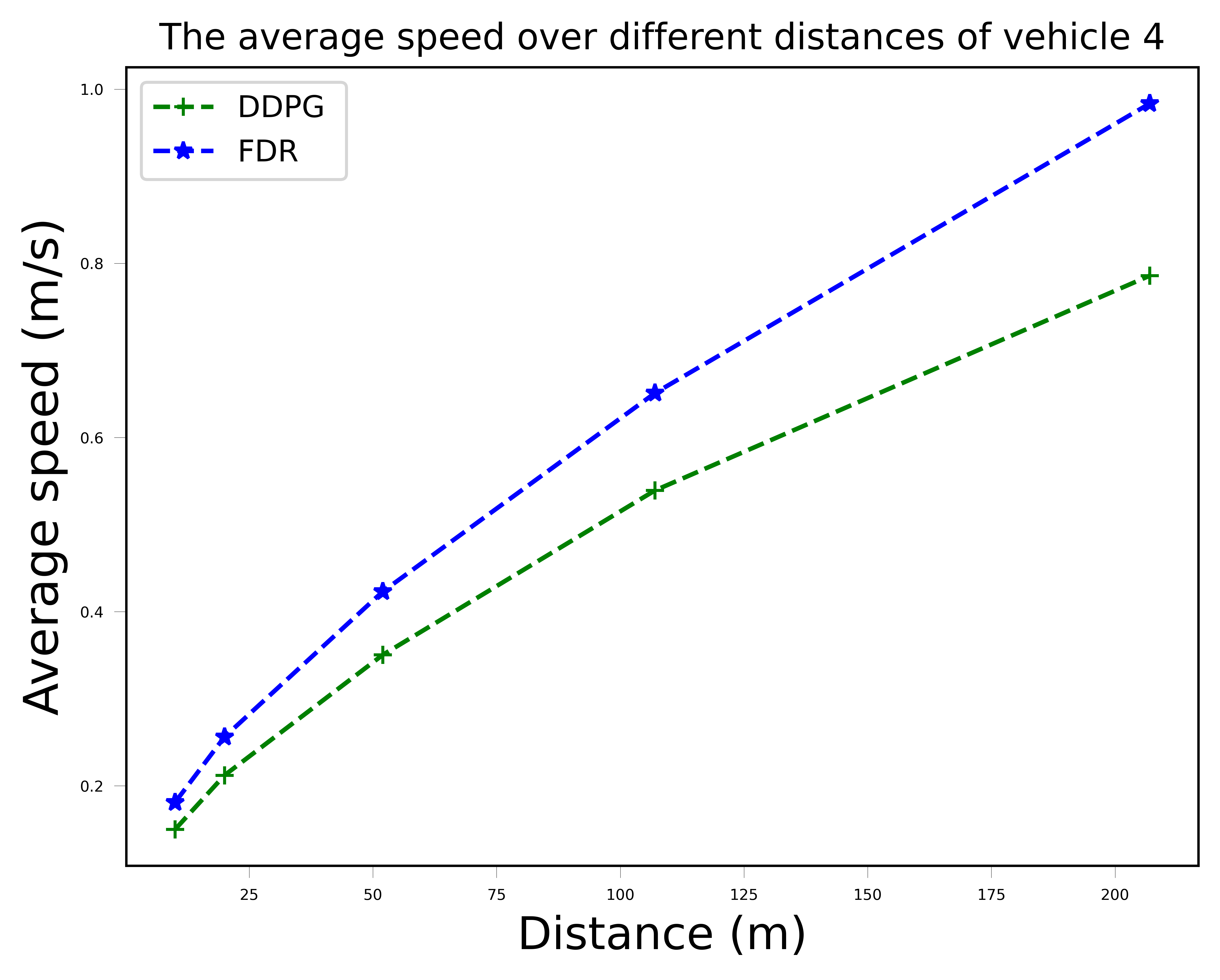}
         \caption{$Vehicle_4$}
         \label{fig:speed4}
     \end{subfigure}
        \caption{The average speed of four vehicles over distances }
        \label{fig:speed}
\end{figure*}

%after Badr
\section{Conclusion}
In this study, our aim was to analyze vehicle control for collision avoidance using Federated Deep Reinforcement Learning (FDRL) techniques, while simultaneously minimizing travel delays, enhancing average vehicle speed, and ensuring safety. To achieve this, we compared the performance of local and global models, namely DDPG and FDDPG, to determine the most effective approach for optimizing vehicle control in collision avoidance scenarios. Through simulations, we observed that the FDDPG model outperformed the DDPG algorithm in terms of controlling vehicles and preventing collisions, all while maintaining data privacy. The FDDPG-based approach exhibited a noticeable reduction in travel delays and an increase in average speed when compared to the DDPG algorithm. These results highlight the potential of FDDPG for practical implementation in intelligent transportation systems. Furthermore, to create realistic environments for evaluating the performance of the deep reinforcement learning algorithms in various traffic setups, we utilized traffic simulation frameworks such as Veins and SUMO.
%In this work, we conducted an analysis of vehicle control for collision avoidance using FDRL techniques, while minimize travel delays, enhance the average speed of vehicles and maintaining safety. We compared the performance of local and global models, DDPG and FDDPG, to determine the most effective method for optimizing vehicle control in collision avoidance scenarios. Simulation results have shown that the FDDPG model outperformed the DDPG algorithm in controlling vehicles and avoiding collisions without sharing sharing sensitive data. The FDDPG-based approach demonstrated a visible reduction in travel delays and an increase in average speed when compared to the DDPG algorithm, highlighting its potential for practical implementation in intelligent transportation systems. Finally, we employed traffic simulation frameworks, namely, Veins and SUMO, to create as realistic as possible environments for evaluating the performance of the DRL algorithms in various traffic setups.

\bibliographystyle{ieeetr}
\bibliography{refs}
\end{document}